# Smart Navigation for In-pipe Robots with Multi-phase Motion Control and Particle Filter

Saber Kazeminasab
*Department of Electrical and Computer Engineering*
*Texas A&M University*
College Station, TX, USA
skazeminasab@tamu.edu

Vahid Janfaza[1]
*Department of Computer Science and Engineering*
*Texas A&M University*
College Station, TX, USA
vahidjanfaza@tamu.edu

Moein Razavi[1]
*Department of Computer Science and Engineering*
*Texas A&M University*
College Station, TX, USA
moeinrazavi@tamu.edu

M. Katherine Banks
*College of Enginnering*
*Texas A&M University*
College Station, TX, USA
k-banks@tamu.edu

**In-pipe robots are promising solutions for condition assessment, leak detection, water quality monitoring in a variety of other tasks in pipeline networks. Smart navigation is an extremely challenging task for these robots as a result of highly uncertain and disturbing environment for operation. Wireless communication to control these robots during operation is not feasible if the pipe material is metal since the radio signals are destroyed in the pipe environment, and hence, this challenge is still unsolved. In this paper, we introduce a method for smart navigation for our previously designed in-pipe robot [1] based on particle filtering and a two-phase motion controller. The robot is given the map of the operation path with a novel approach and the particle filtering determines the straight and non-straight configurations of the pipeline. In the straight paths, the robot follows a linear quadratic regulator (LQR) and proportional-integral-derivative (PID) based controller that stabilizes the robot and tracks a desired velocity. In non-straight paths, the robot follows the trajectory that a motion trajectory generator block plans for the robot. The proposed method is a promising solution for smart navigation without the need for wireless communication and capable of inspecting long distances in water distribution systems.**

I. INTRODUCTION

Water distribution systems (WDS) are one of the strategic infrastructures as they carry water around the globe which is vital to public health. WDS comprises pipeline networks that are prone to incidents that may cause leak or contaminants in water which is harmful to the people's health. Hence, it is required to assess the condition of pipelines frequently to ensure their health. However, these networks are large and buried underground that makes it difficult to access all the parts. To address this issue, in-pipe robots are designed that can perform condition assessment, leak detection, water quality monitoring, visual inspection, and other tasks in pipelines. They move inside pipes with a motion that is independent of water flow. The actuators and electronic parts in these robots require power to operate that is either provided by cables or batteries. Due to limited inspection length of the tethered robots, using battery-powered robots is a more promising solution. The battery-operated robots need to navigate through pipelines with wireless communication with a base station aboveground. It is a challenging task to facilitate a reliable communication link in a harsh environment of water, soil, and rock. In non-metal pipes, it would be possible to use low-frequency radio signals to provide an efficient communication link. However, in some areas, metal pipes are used in distribution networks which destroys radio signals completely and blocks them from broadcasting.

*A. Technical Gap*

Pipelines in WDS are long and have different sizes and configurations. In addition, the environment inside the pipeline have highly-pressured with high-speed flow which is uncertain and disturbing. In the aging pipes, sediments make the inner shape of pipes uneven. It is required that the robots navigate through the long, complicated and uncertain networks with inaccurate maps. Wireless communication that is a common method to this aim is not feasible in **metal** pipes, since they highly attenuate the signals. Also, if the robot operation path is non-straight, a reliable motion control algorithm is needed to move and steer in the network. Hence, smart navigation for in-pipe robots is a challenge for long pipeline inspection.

*B. Literature Review*

The main problem in Autonomous Underwater Vehicles (AUVs) is related to the cyclic problem in Simultaneous Localization and Mapping (SLAM) in autonomous robots, where the robot needs its location to build a map of the environment, and on the other hand, for localization, it needs a map of the environment. For having the best output, the robots should be able to build the map of the environment and localize themselves simultaneously.

The sonar-based localization approach is one of the common ways in the autonomous underwater vehicle. The main problem in solar images is low detectability which is because of some acoustic waves, such as instability and reflection depending on the roughness of the reflector surface. In [2], the authors used sonar-based localization to get sensors' data from the environment to build a coherent map on an unknown environment. They apply a particle filter algorithm to show the vehicle state. Lee et all proposed a method in which by using the sonar images and applying a probability-based framework on them, they can improve the visibility of landmarks and recognize them better [3]. In this method, they applied two probability methods, particle filtering, and Bayesian feature estimation to estimate the feature of objects in the environment. Yuan et all. proposed AEKF-SLAM method which employs the augmented extended Kalman Filter (AEKF) in SLAM to solve the navigation problem in underwater robots [4]. In this method, to get a detailed map, they apply a recursive process to estimate the state parameters and stores the robot poses and landmarks in a single state vector. To overcome the challenges in localization and navigation underwater, the authors in [5] evaluated four different Bayes filter-based algorithms: The Extended Kalman Filter (EKF), Unscented Kalman Filter (UKF), Particle Filter (PF), and Marginalized Particle Filter (MPF) and provided the benefits and drawbacks of each algorithm.

---

[1] Contributed equally.

Robots have great potential for pipe inspection. However, the SLAM problem has been much less widely studied in in-pipe robots [6]. One of the common methods in the localization of in-pipe working robots is landmark-based localization which is mostly used in the navigation of mobile robots. In the pipeline, structures that are distinguishable from the environment can be counted as landmarks such as branches and elbows. Several efforts combine landmark and pose information from sensors to do localization and mapping [7], [8]. In [7], [8], the authors presented a line laser beam projected on the internal surface of the pipeline and deploy the generated unique line pattern to detect landmarks and the pipeline path and perform localization. In [8], they record the information of the robot's position, generated by an oreationation sensor to construct the map of the pipelines. In [7], the authors utilized a line laser beam that has a distinct color and, in this way, the image processing procedure to observe this line is easier. Lee et al. presented a method to do in-pipe robot navigation by landmark recognition method using shadow images obtained by a specially designed illuminator [9]. The proposed method analyzes the images and extracts current position information and also provide the map of the pipeline. Time-of-flight (TOF) camera is a source of 3D imagery for navigation, mapping, and landmark detection for autonomous robot navigation and landmark detection. In [10], the authors propose to utilize the TOF camera for localization and landmark detection. In the proposed method, feature extraction is done by fitting a cylinder to images of the pipeline.

Recently, several efforts have been done on using robots equipped with a camera to solve the autonomous inspection problem in-pipe. Vision sensors such as Fisheye and Omni-directional camera provide a wide field of view and the effectiveness of these sensors in in-pipe inspection and recognition of the environment have been studied in different applications [11]. However, shape measurement is difficult only with the omnidirectional camera. In [11], the authors integrated an omnidirectional rangefinder and 3-D model construction of pipes to solve the self-localization problem of an earthworm robot. They utilize an omni-directional camera and an omni-directional laser to make a rangefinder to construct a piping shape. Diaz et al. developed an adaptive in-pipe inspection robot that can localize the robot and detect and map rust on the pipe [12]. This rust detection method is based on per-pixel classification via image processing. For mapping the rust in the pipe, a checkpoint method is used as a guide for the robot. In [13], the authors described two approaches to increase the localization precision of the pipeline inspection gauge (PIG). First, an inertial navigation system (INS) dynamic model, and second, a 3-D reduced inertial sensor system (RISS).

Utilizing the whisker-like sensors is another method for predicting the direction and also the corner. In [14], the authors used this sensor and developed a bespoke sensor, a competitive sensor package, for corner parameters detection for autonomous navigation in an unknown pipe network. In [15], the authors proposed a method to use the advantages of using an acoustic signal with pose-graph optimization to improve the estimation of a robot's trajectory. They utilized four methods to construct the pose graph from the spatially varying hydrophone acoustic signal.

One of the main problems in SLAM is that pipes tend to be relatively featureless making robot localization a challenging problem. Recent efforts have been done to reduce the systematic noise and observation noise which can result in an inaccurate judgment for the mobile robot. KF (Kalman Filter) can use in the linear motion system to eliminates the Gaussian noise. EKF (Extended Kalman Filter) is effective in a nonlinear system that simulates the linear system with Taylor's first order expansion. PF (Particle Filter) can express the posterior probability distribution of the systematic status and is useful in the nonlinear none Gaussian system [16]. Ke Ma et all. proposed PipeSLAM for simultaneous localization and mapping in metal water pipes which used RaoBlackwellised particle filter (RBPF) for estimation [6]. This method generates a map of pipe vibration amplitude over space by exciting pipe vibration with a hydrophone. Then, it utilizes this new type of map for the SLAM algorithm. In [15], the authors created a new type of map using signals from hydrophone excitation of the metal pipe. They applied a signal alignment and averaging algorithm based on dynamic time warping to address the problem of spatially calibrating the map. For localization problem, they utilize PF and also EKF. YuPei Yan et all. utilized PF-SLAM algorithm which combines the Particle Filter and FastSLAM algorithm for obtaining the accurate moving direction and also navigation of mobile robot [17]. They investigated a visual simultaneous localization and mapping (VSLAM) method for locating a pipe inspection robot and generating a map of the environment. In this method, they integrate the information of an inertial navigation sensor, ultrasonic rangefinders sensors, and multiple CCD cameras.

*C. Our Contribution*

In this paper, we introduce a method that omits the need for communication with base station for navigation. The method is combined with a two-phase motion control algorithm to inspect large distances of distribution network long inspection and smart navigation in WDS.

*D. Paper Organization*

In section II, a description of the in-pipe robot is provided. In section III, a multi-phase motion controller is described that facilitates reliable motion in straight and non-straight paths. In section IV, the particle filtering method is introduced. The combination of particle filtering and multi-phase motion control is presented in section V. The paper is then concluded in section VI.

## II. OUR IN-PIPE ROBOT

Our proposed in-pipe robot is in class of underactuated mobile robots that have adjustable size. It is classified as wall press wheeled robots that press the pipe wall to stabilize themselves and crawl in the direction of pipe axis [1]. Using underactuation mechanism, helps to control more degrees of freedom of the robot with fewer actuators. The robot is designed and manufactured in a way that is not a source of toxic materials for water. To this aim, a seal mechanism has been designed for all sensitive components. The robot consists of three different modules: 1- Central processor. 2- Adjustable arm module. 3- Sealed actuator module.

The central processor of the robot hosts the motion control unit, sensor module unit to take measurements, and power supply module. The control unit controls the motion of the robot in pipe in a way that is stable and robust to uncertainties and disturbances. The sensor module unit is consists of sensors for measuring a parameter in water (i.e. quality monitoring) or pipe environment (i.e. crack and leak). The power supply (i.e. battery) provides the power needed for the robot and is located in the central processor. The power is transformed to the internal space of the central processor via wire.

To make the robot size adjustable, three adjustable arms are used. Each arm comprises an arm, a passive spring and a pair of ball bearings. The arms are connected to the central processor with 120 degrees angle with ball bearings. The passive spring is anchored on the arm and the central processor. At the end of the each arm, an actuator module is located that makes the robot move inside pipe. The actuator module comprises a gear motor equipped with an incremental encoder and a wheel that is mounted on the arm with a pair of ball bearings. Fig 1 shows the overall view of our in-pipe robot and different parts of it. The details of fabrication and prototyping can be found in [1].

*A. Modelling*

Fig. 2 shows the robot in a pipe that has an inclination angle with respect to the horizontal surface. The traction forces of the wheels are specified as $F_1$, $F_2$, and $F_3$ generated by motors. The robot does not have linear motion in y-axis and z-axis. Also, the rotation around x-axis is zero.

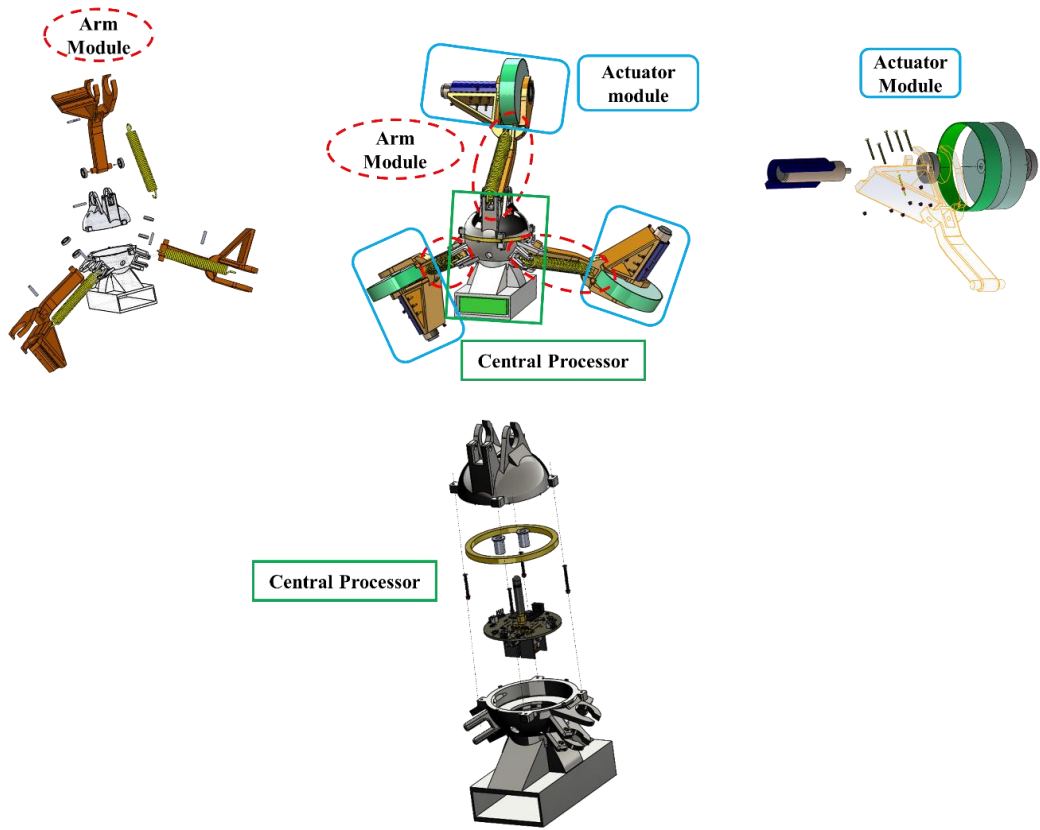

Fig. 1. Our proposed in-pipe robot and its components.

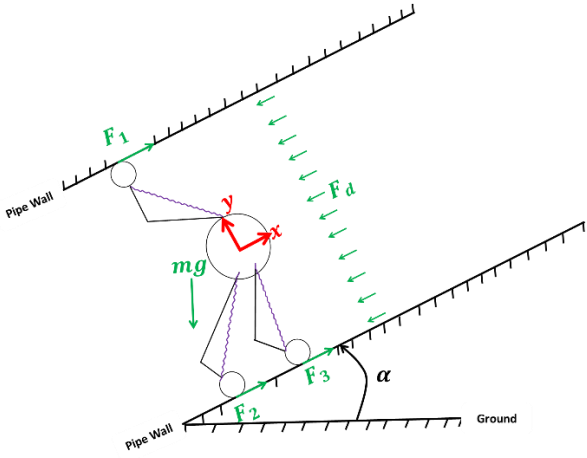

Fig. 2. Free body diagram of the robot in a pipe with an inclination angle, $\alpha$.

In the x-axis (pipe axis), we have:
$$\sum F_x = ma \rightarrow F_1 + F_1 + F_1 - mg\sin(\alpha) - F_d = ma = m\ddot{x} \quad (1)$$

In Eq. (1), $m$, $g$, $\alpha$, and $F_d$ are the robot mass, gravity force, inclination angle, and drag force, respectively. The drag force is the result of the difference between the robot and water flow velocity [18]. The resultant torques around y-axis considering point $O$ in Fig. 2 are as follows:
$$\sum M_Y = I_{yy}\ddot{\phi} \rightarrow \frac{\sqrt{3}}{2}F_3L\cos(\theta_3) - \frac{\sqrt{3}}{2}F_2L\cos(\theta_2) = I_{yy}\ddot{\phi} \quad (2)$$
where $I_{yy}$ and $\phi$ are the robot mass moment of inertia in y-axis and rotation around y-axis, respectively. $L$ is the arm length and $\theta_i, i = 1,2,3$ are arm angles that are shown in Fig. 2. Also, by doing the dynamic force analysis around z-axis, we have:
$$\sum M_z = I_{zz}\ddot{\psi} \rightarrow \frac{1}{2}F_3L\cos(\theta_3) + \frac{\sqrt{3}}{2}F_2L\cos(\theta_2) -$$
$$F_1L\cos(\theta_1) - mg\cos(\alpha)L\sin(\theta_1) = I_{zz}\ddot{\psi} \quad (3)$$
Similar to Eq. (2), $I_{zz}$ and $\psi$ in Eq. (3) are the robot mass moment of inertia in z-axis and rotation around z-axis, respectively.

### III. MULTI-PHASE MOTION CONTROL

The pipelines include complicated structures; straight and non-straight configurations. The robot is supposed to move inside these configurations. In this section, we design a multi-phase motion controller that enables the robot to pass through straight and non-straight configurations like bends and T-junctions. In the straight path, first the robot needs to stabilize itself in pipe, and track a desired velocity. To this aim, we develop a linear quadratic regulator (LQR) controller that is a full state feedback controller and a good solution for stabilizing control problems [19]. Therefore, we defined $x = [\phi \quad \dot{\phi} \quad \psi \quad \dot{\psi}]$ as stabilizing states and derived the system auxiliary matrices by linearizing Eqs. (2) and (3) around the equilibrium point, $x_e = [0 \quad 0 \quad 0 \quad 0]$ [19], [20]. We calculate the controller gain metric, $k$, with the system matrices, and the output control signal, $u$, is calculated as:

$$u = -kx \quad (4)$$

With this control signal, the robot stabilizes itself during operation. The robot also needs to track a desired velocity in pipelines. To this aim, we validated that if the angular velocity of the wheels are approximately equal, the linear velocity of the robot, $v$ is calculated as [20]:
$$v = 2\pi R\omega \quad (5)$$
Where $R$ is wheel radius and $\omega$ is wheel angular velocity. To control the velocity of the robot, the desired velocity of the wheels are calculated considering the desired velocity of the robot and Eq. (5). Three proportional-integral-derivative (PID) controllers are tuned to control the desired angular velocity of each wheel. We combined the LQR and PID controllers to shape the stabilizer-velocity tracking controller. The performance of this controller is analyzed with experimental results [20]. The experiment is repeated for four iterations and the results are shown in Fig. 3. The controller is able to stabilize the initial deviations of $\phi$ and $\psi$ within two seconds of the initiation of movement. Also, it is able to track the desired velocities in the iterations (see Fig. 3). (This Figure must be bigger and in one column layout; low quality)

As mentioned, the robot should be able to navigate non-straight configurations. To this aim, we design a novel controller algorithm for the robot, based on Variable Velocity Allocation (VVA). In this method, the robot allocates different desired angular velocities for the wheels based on the configuration type. Similar to the LQR-PID based controller, three PID controllers, track the desired velocities. An error check module is also developed that measures the rotation angle of the robot continuously in the junction. After completion of the rotation, the controller would be deactivated. In Fig. 4, the controller algorithm for non-straight paths is shown.

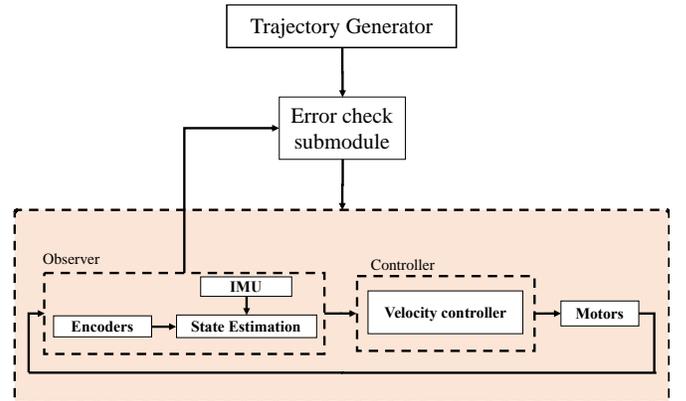

Fig. 4. The controller algorithm for the robot that steers it in non-straight configurations to the desired paths. The trajectory generator block defines the desired angular velocities for each wheel. The error check submodule checks the rotation angle of the robot in the junction.

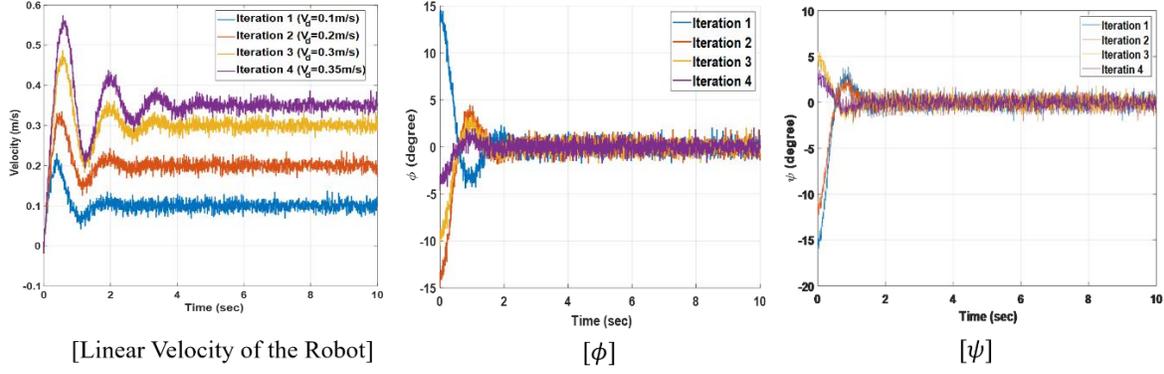

[Linear Velocity of the Robot]  [$\phi$]  [$\psi$]

Fig. 3. The experimental results to analyze the performance of LQR-PID based controller in four iterations. The controller cancels the initial deviations of stabilizing states, and reach to a desired velocity. Source: Adapted from [20].

To evaluate the functionality of the controller in non-straight paths, we modeled the robot in two common non-straight paths; bends and T-junctions in ADAMS software. Also, the controller is implemented in MATLAB Simulink toolbox and linked with the dynamical model in ADAMS. The model along with the control plant is co-simulated and the results are shown in Fig. 5. In this figure, the sequences of motion show that the robot with the controller has a reliable motion in non-strait pipe path.

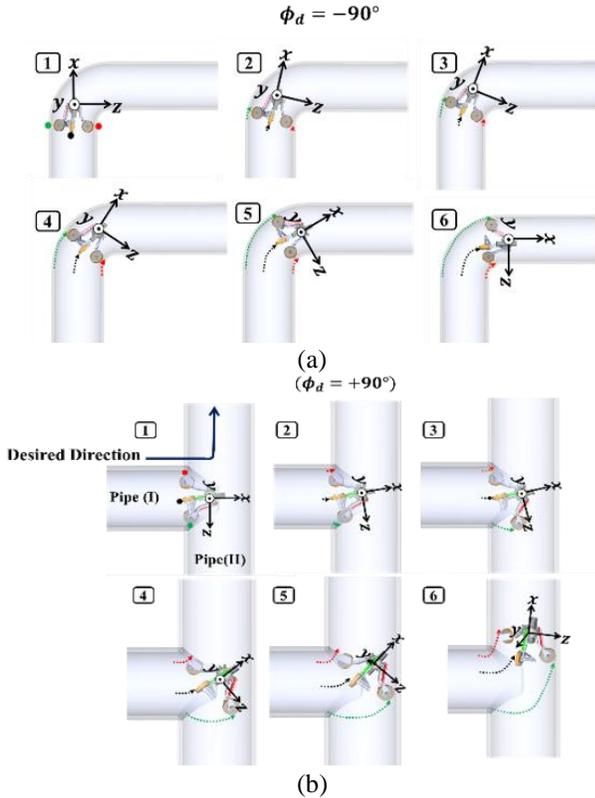

Fig. 5. The performance of the robot and the controller in non-straight paths in ADAM-MATLAB co-simulation. A coordinate system is considered on the central processor. $\phi_d$ is the desired rotation in each junction. a) The sequences of motion of the robot in a bend. b) The sequences of motion of the robot in a T junction.

So far, we designed a robot and a two-phase motion controller that stabilizes the robot in straight paths and steers it in non-straight paths. The problem here is that the robot should be "aware" of when to choose the appropriate motion controller. As we mentioned, wireless communication in metal pipes is not feasible. So, we propose a positioning technique based "particle filtering" that determines the location of the robot based on measurements of pipe environment, and combine it with a novel method to facilitate smart navigation for the robot.

## IV. PARTICLE FILTERING METHOD

The particle filtering method is a numerical algorithm to estimate some unknown parameter(s) of a system that cannot be measured directly. To do so, this method uses the relationship of directly measured parameter(s) of the system with the unknown parameter. First, a random sample of points (particles) with some weights are assigned to the unknown parameter(s) (e.g., using a random uniform distribution). Next, the weights of those initial particles will be updated based on the measured parameters and their relationship with the unknown parameter(s). Then, with respect to the updated weights, the particles are resampled. Finally, using a model, the resampled particles would be propagated in time. By repeating these steps over time, the particles will converge to the true estimate of the unknown parameter(s). The superiority of the particle filtering method over the Kalman filter method is that, in the Kalman filter, the distribution of the measured parameter(s) and the estimated distribution of the unknown parameter(s) must be both Gaussian, while the particle filtering method does not have this requirement [21]. This feature of the particle filtering method along with the capability of estimation is useful in localization and navigation in uncertain environments.

Using the particle filtering method, the position of the mobile robot can be estimated by a sensor that can detect the nearby objects along with a measurement of how the robot is moving. In Algorithm 1, we present how to use particle filtering method for localization and mapping. For this purpose, first, we uniformly randomize both the position and orientation of the robot across the entire map $< x_{t-1}^j, w_{t-1}^j >$; the higher number of random points leads to higher accuracy but lower convergence speed. Then using the information from the sensor measurements ($z_t$), we update particles and give higher weight to the more probable locations for the robot and remove the points that are unlikely to be the current position of the robot. With the new weights, we resample the points using Adaptive Monte Carlo Localization (AMCL). Next, we add random noise to the information from how the robot is moving and apply that noisy movement ($u_t$) to every point in the new sample. $\{S_{t-1} = <x_{t-1}^j, w_{t-1}^j>, u_t, z_t\}$ is the set of particles in which $<x_{t-1}^j, w_{t-1}^j>$ is discrete sample distribution of positions ($x$) with weights ($w$), $u$ is the new known action or the movement of the robot and $z$ is new measurement from the directly measured parameter(s).

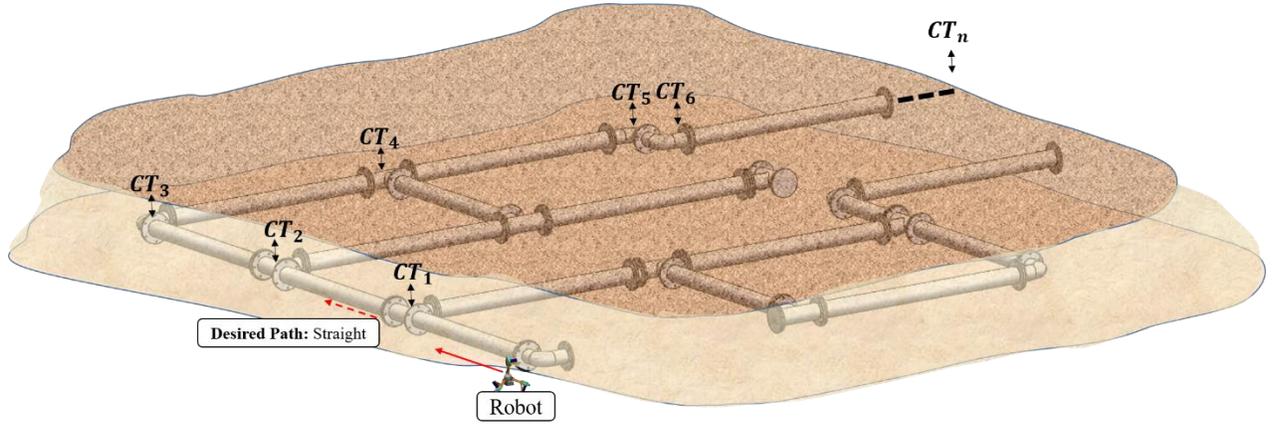

Fig. 6. The proposed method to provide the robot with the map of operation in the pipeline network. $CT_i$ ($i = 1,2,\dots,n$) contains the configuration type and the desired path in the non-straight paths. Based on this method, the robot distinguishes its way in the complicated pipeline network.

**Algorithm 1: Particle Filtering Method**

| | |
|---|---|
| 1 | $S_t = \emptyset, \eta = 0$ |
| 2 | **for** $i$ from 1 to $n$ **do** |
| 3 | Sample index $j(i)$ from the discrete distribution given by $w_{t-1}$ |
| 4 | Sample $x_t^i$ from $p(x_t|x_{t-1}, u_t)$ using $x_{t-1}^{j(i)}$ and $u_t$ |
| 5 | $w_t^i \leftarrow p(z_t|x_t^i)$      *Recompute the weights by likelihood* |
| 6 | $\eta \leftarrow \eta + w_t^i$      *Update normalization factor* |
| 7 | $S_t \leftarrow S_t \cup \{<x_t^i, w_t^i>\}$ *Add the updated particle to the set* |
| 8 | **end for** |
| 9 | **for** $i$ from 1 to $n$ **do** |
| 10 | $w_t^i \leftarrow w_t^i/\eta$ |
| 11 | **end for** |

## V. PROPOSED SMART NAVIGATION METHOD USING PARTICLE FILTERING AND MUILTI-PHASE CONTROLLER

So for, we have a flexible in-pipe robot and a multi-phase motion controller. The controller enables the robot to have stabilized motion in straight paths in one phase and also steers the robot in non-straight paths in another phase. We mentioned that particle filtering method is an appropriate option for navigation when the robot has a **map** of operation while it does not know about its exact location in the map. In this part, we propose a method that enables the robot to localize itself in the pipe and also, switches between different phases of motion control.

We provide the robot with the map of the pipeline by giving it an array which includes the non-straight configuration types of the pipeline that the robot operates in. The array is represented as:

$$\mathbf{CT}=[CT_1 \quad CT_2 \quad \dots \quad CT_n] \qquad (6)$$

where $CT_i, i = 1,2,\dots,n$ defines the configuration type of each non-straight configuration and the desired new path. For example, in Fig. 6, the configuration type that is associated with $CT_1$ is a T-junction and the desired path for the robot in this configuration is straight. For localization purpose, an ultrasonic sensor is used to provide information about the robot's surrounding environment. It continuously sends sonar waves to the pipe (Fig. 7). In other words, the ultrasonic sensor takes the measurements of the environment at each time step that we defined it as $z_i$ in the previous section. With the information that is stored in **CT** matrix and the sensor measurements, the robot is able to detect the non-straight configurations. So far, the robot is headed in a straight path and with the provided map of the pipeline, a sonar sensor, and particle filtering method, it defines the configuration type of non-straight paths. In other words, the robot first starts moving in a straight path, then reaches to a non-straight path and with the proposed method (i.e., using sonar sensor and particle filtering) the position of the robot will be determined. As for the motion control algorithm, the robot motion is controlled with the stabilizer-velocity tracking controller. In the non-straight path, it follows a motion controller algorithm we explained in Fig. 4. The robot in this motion control, passes through the non-straight configuration and then needs to follow a straight path in the pipeline. Hence, it needs to switch to stabilizer-velocity tracking controller. Switching between two motion controllers is done with the particle filtering method. The robot detects the non-straight path and the configuration type by continuously monitoring the pipe and then switches to the other phase of the motion controller. Fig. 8 shows the synchronized two-phase motion controller and the particle filtering method.

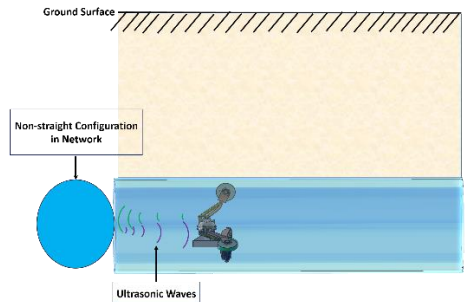

Fig. 7. Using sonar sensor to detect the non-straight configurations and their type (e.g., bend, T-junction).

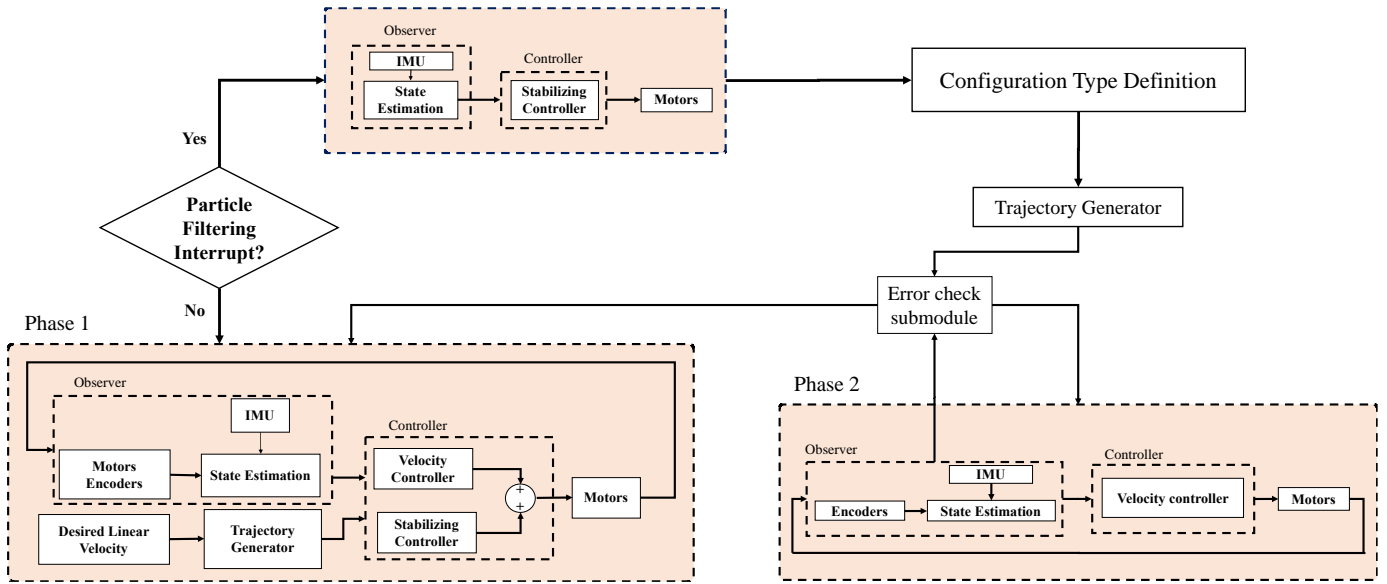

Fig. 8. The algorithm for smart navigation for the robot with the particle filtering method (in metal pipes).

*A. Experiments*

We used an experiment to evaluate the performance of the proposed synchronized particle filter method and the control algorithm. For that, we attached a HC-SR04 ultrasonic sensor module in front of the robot and placed the robot in a 14-in diameter pipe. The robot started moving with stabilizer-velocity tracking controller (phase 1) with the speed of 10 cm/s. It moved towards the end of the pipe and when it reached to the distance of around 14-in from the end of the pipe, it entered to the phase 2 of the controller and stopped with stabilized configuration. Hence, the results show that the proposed localization and navigation method gives the robot the potential for smart navigation.

## VI. Conclusion

Navigation for in-pipe robots is a challenging task, especially in metal pipes and the pipeline networks with complicated configurations. Wireless communication systems are not feasible as radio signals are destroyed in metal environments. In this paper, we addressed this challenge by designing a two-phase motion controller for our previously designed in-pipe robot and particle filtering method. In phase one of the controller, a stabilizer-velocity tracking controller stabilizes the robot in straight paths and enables the robot to track a desired velocity. Also, in phase two (of the controller), the robot changes its direction to the desired path in non-straight paths like bends, T-junctions etc. The switching between two phases of the controller are performed with the particle filtering method that determines the non-straight configurations with an ultrasonic sensor. The map of the robot operation procedure is given to the robot's firmware with a novel approach that the particle filtering uses for configuration type determination. Performance of the motion controller algorithm is validated with simulation and experimental results. The proposed method is a promising solution for smart navigation in in-pipe robot field.

In future, we plan to expand the smart navigation algorithm to enable the robot to perform desired tasks such as water quality monitoring (timing of the robot), and also test its performance with experiments.

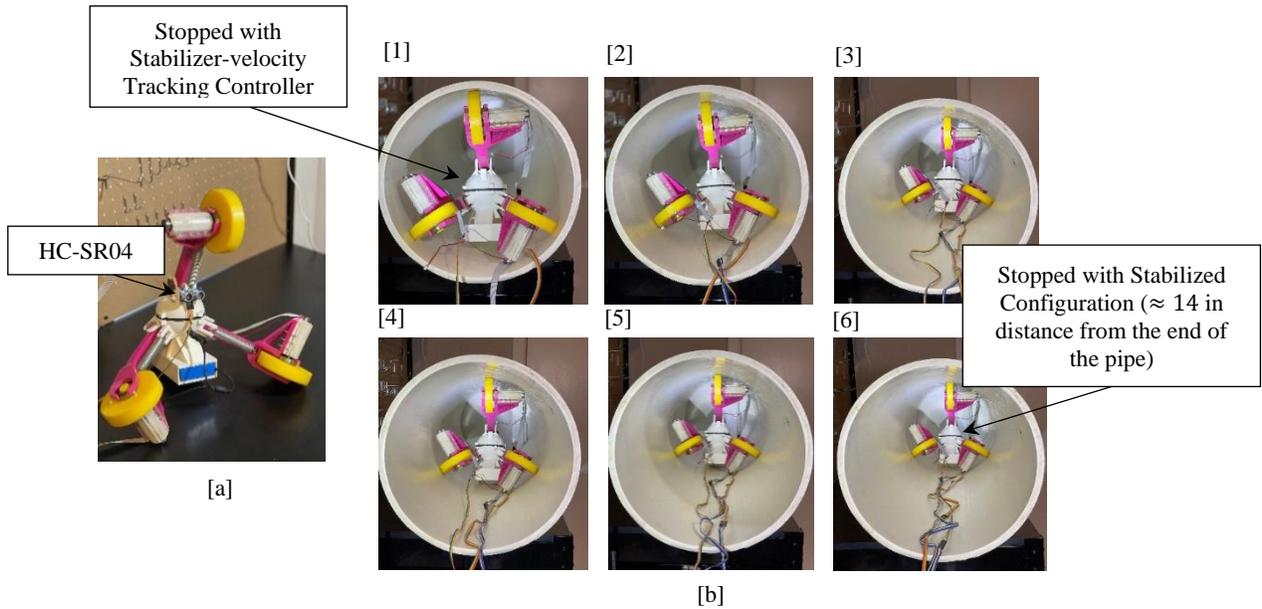

Fig. 9. [a] Prototype of the Robot Equipped with HC-SR04 Ultrasonic Sensor. [b] Sequences of motion of the robot in 14-in PVC pipe. The robot starts with the stabilizer-velocity tracking controller and switches to the stabilizer controller when its distance from the front obstacle is around 14-in.